\documentclass[runningheads]{llncs}

\usepackage{graphicx}
\usepackage{todonotes}
\usepackage{amsmath}
\usepackage{amssymb}
\usepackage{fancyref}
\usepackage{subfig}
\usepackage{booktabs}
\usepackage{multirow}
\usepackage{pgfplots}
\usepackage{wrapfig}
\usepgfplotslibrary{statistics}
\newcommand{\ra}[1]{\renewcommand{\arraystretch}{#1}}

\newcommand{\printfnsymbol}[1]{%
  \textsuperscript{\@*}%
}

\begin{document}
\title{3DQ: Compact Quantized Neural Networks for Volumetric Whole Brain Segmentation}

\titlerunning{3DQ: Compact Quantized Neural Networks}

\author{Magdalini Paschali\inst{1}\thanks{The authors contributed equally.} \and Stefano Gasperini\inst{1}\printfnsymbol{1} \and
Abhijit Guha Roy\inst{2} \and \\
Michael Y.-S. Fang\inst{3}\and
Nassir Navab\inst{1,4}
}
\authorrunning{M. Paschali, S. Gasperini et al.}

%index{Paschali, Magdalini}
%index{Gasperini, Stefano}
%index{Guha-Roy, Abhijit}
%index{Fang, Michael Y.-S.}
%index{Navab, Nassir}

%tocauthor{Magdalini Paschali, Stefano Gasperini, Abhijit Guha-Roy, Michael Y.-S., Nassir Navab}

\institute{
Computer Aided Medical Procedures, Technical University of Munich, Germany
\and 
Department of Child and Adolescent Psychiatry, Psychosomatic and Psychotherapy, Ludwig Maximilian University, Munich, Germany
\and
Department of Physics, University of California Berkeley, USA
\and
Computer Aided Medical Procedures, Johns Hopkins University, USA
}
\maketitle              % typeset the header of the contribution
\begin{abstract}

Model architectures have been dramatically increasing in size, improving performance at the cost of resource requirements. In this paper we propose 3DQ, a ternary quantization method, applied for the first time to 3D Fully Convolutional Neural Networks (F-CNNs), enabling 16x model compression while maintaining performance on par with full precision models. We extensively evaluate 3DQ on two datasets for the challenging task of whole brain segmentation. Additionally, we showcase our method's ability to generalize on two common 3D architectures, namely 3D U-Net and V-Net. Outperforming a variety of baselines, the proposed method is capable of compressing large 3D models to a few MBytes, alleviating the storage needs in space-critical applications.

\end{abstract}

\section{Introduction}

Fully Convolutional Neural Networks (F-CNNs) have been incorporated successfully into numerous Computer Assisted Diagnosis (CAD) systems performing various medical image analysis tasks with increasing difficulty and requirements. Hence, their size has grown drastically reaching commonly hundreds of layers and several million parameters.

An additional factor towards the size explosion of F-CNNs in CADs is that medical data is in most cases volumetric by nature and has been continuously increasing in resolution.
This growth of F-CNNs has shortcomings regarding resource requirements, such as computation, energy consumption and storage. 
Towards this end, we propose, for the first time in 3D F-CNNs, a quantization-based technique that offers model compression up to 16 times without any loss in performance.

To this day, various segmentation methods process volumetric data per slice using 2D F-CNNs~\cite{quicknat}. Such methods yield satisfying results but are not able to fully exploit the contextual information from adjacent slices. Existing 3D F-CNNs, like V-Net~\cite{milletari2016vnet}, 3D U-Net~\cite{cciccek20163dunet} and VoxResNet~\cite{voxresnet} have achieved state-of-the-art performance in a plethora of segmentation tasks utilizing 3D kernels. However, their millions of parameters require a large amount of space for storage.

Deep learning-based CADs deploying 3D F-CNNs are being integrated into the medical workflow~\cite{hospital}. Hospitals, which were already burdened with storing a myriad of large medical records, now have to allocate additional storage for the trained models in CADs.
On top of that, the emerging field of patient-specific care~\cite{vivanti2018patient}, while improving personalized diagnosis and monitoring, will increase the need for storage in medical facilities even further.

Model compression has been an active area of research in the past few years aiming at deploying state-of-the-art F-CNNs in low-power and resource-limited devices, like smartphones and embedded electronics.
An additional potential use-case is represented by the decentralized training approach of Federated Learning~\cite{konevcny2016federated}. Despite maintaining the privacy of client data, iteratively sending over the internet millions of parameters to train a global model becomes unstable with unreliable connections, and compressed models would improve the process.

Multiple techniques, like parameter pruning, low-rank factorization, knowledge distillation and weight quantization~\cite{compressionsurvey}, have been proposed to compress the size of F-CNNs without compromising their performance. Specifically, weight quantization to binary~\cite{rastegari2016xnor} and ternary values~\cite{heinrich2018ternarynet},~\cite{zhu2016ttq},~\cite{lin2018defensivequantization} has been among the most popular methods for this task. This can be attributed to its additional advantage of allowing for impressive speed-up during training and inference by approximating convolutions with XNOR and bitcounting operations~\cite{rastegari2016xnor}. Even though it is generally accepted that XNOR-Net revolutionized this field~\cite{rastegari2016xnor}, there is a significant trade-off between performance and speed, not ideal for medical applications. TernaryNet~\cite{heinrich2018ternarynet} was the first attempt in medical imaging to create compact and efficient F-CNNs utilizing ternary weights, where a 2D U-Net was employed for the task of per-slice pancreas CT segmentation. However, extending quantization to 3D F-CNNs has yet to be explored.

In this paper, our contribution is two-fold: 1) We propose, for the first time in 3D F-CNNs, a quantization mechanism with a novel bit-scaling scheme which we name 3DQ. 3DQ integrates two trainable scaling factors and a normalization parameter that increases the learning capacity of a model while maintaining compression. 2) We extensively evaluate 3DQ on the challenging task of 3D whole brain segmentation, showcasing that state-of-the-art performance can be combined with impressive compression rates.

\section{Method}

\subsection{Weight Quantization}
The main goal of our quantization method is approximating the full precision weights $W$ by their ternary counterparts \{-1, 0, 1\}, $\tilde W$, as formalized in Eq.~\ref{eq:quant}. 

\begin{figure}[t]
\centering
  \includegraphics[width=0.88\textwidth]{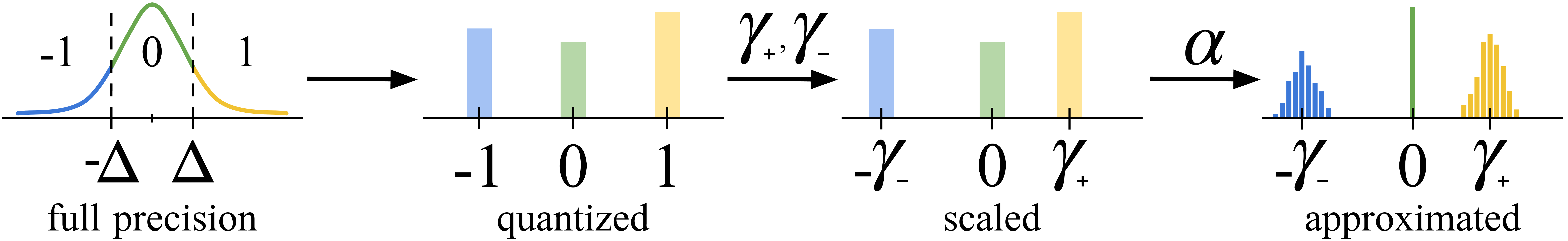}
   \caption{Overview of 3DQ. The full precision weights are quantized into ternary values, scaled by $\gamma_{\pm}$ and then further diversified by the factor $\alpha$.}
   \label{fig:quantization}
\end{figure}

The first step is computing the threshold $\Delta$, based on which $W$ will be assigned into three quantization bins. Various approaches leverage a single $\Delta$ for the entire network~\cite{heinrich2018ternarynet}. However, 3DQ computes one $\Delta$ per layer $l$, to maintain the variability in weight range values within each layer and avoid weight sparsity~\cite{zhu2016ttq}. Specifically, each $\Delta_l$ is computed as $\Delta_l = t \cdot \max{(|W_l|)}$: the maximum absolute value of the weights in each layer is multiplied by a constant factor $t$ which moderates the weight sparsity and is consistent among all layers. We set $t$ to $0.05$, as suggested by~\cite{zhu2016ttq} as it proved to be the optimal trade-off between sparsity and accuracy.

After thresholding, the acquired ternary weights $\tilde W$ are multiplied by a set of scaling factors, since training an entire F-CNN with \{-1, 0, 1\} weight values would lead to substantial suboptimal performance. 3DQ utilizes two scaling factors, $\gamma^{+}_{l}$ and $\gamma^{-}_{l}$~\cite{zhu2016ttq}, which are variables learned for each layer $l$ during training, in contrast to previous methods such as~\cite{heinrich2018ternarynet,rastegari2016xnor}.
%\vspace{-0.1cm}
\begin{align}
    W_l \approx \tilde W_l =
    \begin{cases}
        + \gamma_l^{+} \cdot \alpha & \text{if ~} W_l > \Delta_l\\
        0 & \text{if ~} |W_l| < \Delta_l\\
        - \gamma_l^{-} \cdot \alpha & \text{otherwise}.
    \end{cases}
    \label{eq:quant}
\end{align}

Furthermore, unlike~\cite{zhu2016ttq}, we incorporated an additional scaling factor $\alpha$ into 3DQ~\cite{ttw}. $\alpha$ is calculated from $W$ as the average of the weights with an absolute value larger than $\Delta_l$, as $\alpha = \frac{1}{n_{\Delta_l}}\sum |\tilde W_l| |W_l|$, where $n_{\Delta_l}=\sum|\tilde W_l|$. $\alpha$ enhances the approximation of the full precision weights, since it spreads the quantized values within the same bin, increasing the expressivity and diversity of the weights in between the various channels of each layer. The quantization process is schematically summarized in Fig.~\ref{fig:quantization}.

\subsection {Compact Weight Storing}

As explained above, both $W$ and $\tilde W$ are required during training in order to achieve the optimization of the model weights and the learned scaling factors $\gamma^{+}$ and $\gamma^{-}$.
However, during inference, the full precision weights are no longer necessary and therefore there is no need to store them. 

\begin{figure}[t]
\centering
  \includegraphics[width=0.88\textwidth]{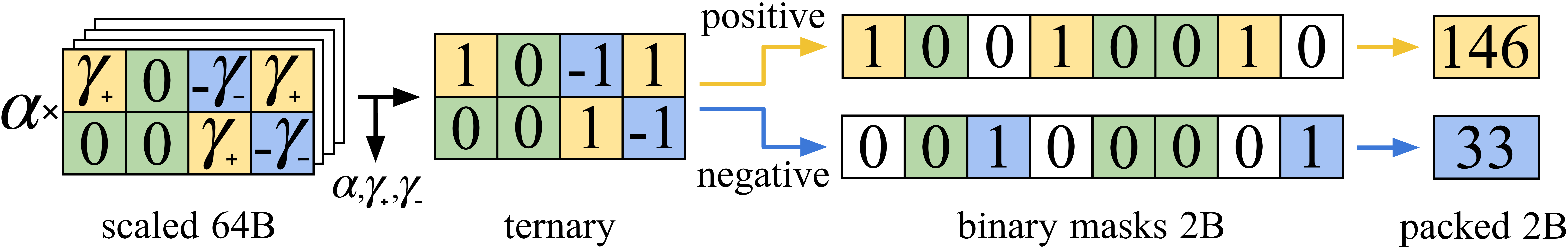}
   \caption{Overview of 3DQ compression method to store the model parameters. The scaling factors are separated from the ternary values, which are further split into binary vectors, then packed from 8 bits to a byte.}
   \label{fig:compression}
\end{figure}

Since after the scaling of $\tilde W$, the values still use 32 bits, it is important to store the model ensuring the 16x compression rate offered by the ternary weights. Towards this end, we separate each kernel into three components, as can be seen in Fig.~\ref{fig:compression}: 1) The trained scaling factors $\gamma^{+}$ and $\gamma^{-}$, which are as many as the layers of the chosen F-CNN architecture. 2) The $\alpha$ values, which are computed from the full precision weights and sum up to as many as the channels at each layer. 3) The ternary weights, which constitute the majority of the model parameters, summing up to millions of values in cases of 3D models. 

The scaling factors are stored as full precision variables, requiring 32 bits of disk space per value. Meanwhile, each ternary weight kernel is split into two binary masks, one for the positive weights and one for the negatives. The unmarked locations in both masks stand for the zero weights. Afterwards, the masks go through bit packing and 8 weight bits are stored in 1 byte. The same process is followed backwards to restore the saved models: first unpack the weight values, then multiply them with the saved scaling factors. This method achieves impressive compression rates, particularly crucial for large 3D models, where oftentimes 45M parameters need to be stored for each network~\cite{milletari2016vnet}.

\section{Experimental Setup}

\noindent\textbf{Datasets} We evaluated 3DQ on two challenging and publicly available medical imaging 3D segmentation datasets, namely the Multi-Atlas Labelling Challenge (MALC)~\cite{malcoasis} and the Hippocampus (HC) Segmentation dataset from the Medical Decathlon challenge~\cite{medicaldecathlon}. MALC is part of the OASIS dataset and contains 30 whole brain MRI T1 scans with manual annotations. The input volumes are sized $256\times256\times256$, which were sampled in cubic patches of size 64.
Maintaining the original challenge split, we used 15 scans for training and 15 for testing. We considered 28 classes for the segmentation, following~\cite{quicknat}, and we repeated all the experiments 5 times.

HC includes 263 training samples sized on average $36\times50\times35$, which we padded to cubes sized 64.
%$64\times64\times64$.
Due to the public unavailability of the test set, we performed 5-fold cross-validation for all our experiments, dividing the given dataset to 80/20 patient-level splits. In this dataset, the volumes are segmented into 3 classes, 2 parts of the hippocampus (hippocampus proper and hippocampal formation) and the background~\cite{medicaldecathlon}.
\par

\begin{figure}[t]
\centering
  \includegraphics[width=0.80\textwidth]{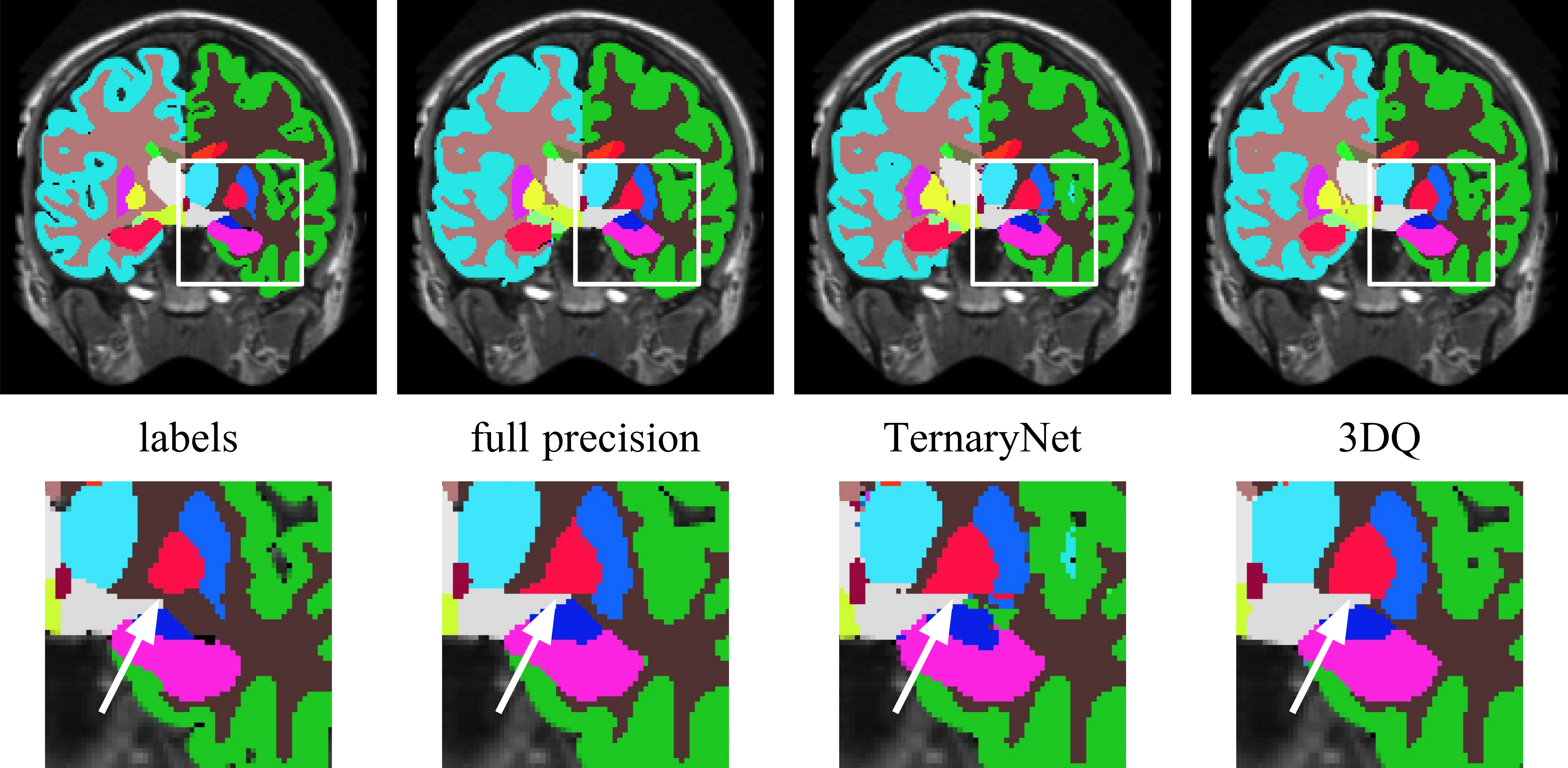}
   \caption{Qualitative results of 3DQ compared with baseline methods. White arrows on the zoomed views show the superior segmentation performance of 3DQ.}
   \label{fig:segmentation}
   \vspace{-0.1cm}
\end{figure}

\noindent\textbf{Model Training}
To highlight the generalizability of our method, for the first time, we quantized common 3D F-CNN architectures, namely 3D U-Net~\cite{cciccek20163dunet} on MALC and HC, and V-Net~\cite{milletari2016vnet} on MALC. The aforementioned models are highly suitable candidates for quantization and compression, since they have 16M and 45M trainable parameters respectively and require up to 175MB of storage space.

For both datasets, we trained the models with a composite equally-balanced loss function comprised of Dice loss and weighted cross entropy. The weights were computed with median frequency balancing~\cite{roy2017error} to circumvent class imbalance. We used an Adam optimizer, initialized with learning rate 0.0001 for 3D U-Net and 0.00005 for V-Net. We trained all models on an NVIDIA Titan Xp Pascal GPU and implemented 3DQ on PyTorch. Although all 3D U-Net models were trained from scratch, we found beneficial starting the quantized experiments on V-Net from a pretrained version.

\noindent \textbf{Evaluation Metrics}
The main idea of our approach is compressing quantized models, without sacrificing performance with respect to their full precision counterparts. Therefore, the evaluation, which we will discuss in Section~\ref{sec:results}, is based on two different criteria: the Dice coefficient achieved by the models across volumes and the storage space required to save them.

\noindent \textbf{Ablative testing}
In order to showcase the effectiveness of the main components of 3DQ, namely the ternary weights and the addition of the scaling factor $\alpha$, we performed ablative testing. Comparing 3DQ with BTQ, an adjusted binarized version of Trained Ternary Quantization (TTQ)~\cite{zhu2016ttq},
%its binary-quantized counterpart, 
is an essential experiment to highlight the benefits of ternary weights. Furthermore, we compared 3DQ against TTQ to demonstrate the contribution of $\alpha$ as a scaling factor.

\noindent \textbf{Baseline comparison}
Furthermore, 3DQ was evaluated against its full precision counterpart in order to investigate whether quantized models are able to match the performance of full precision networks. 3DQ was also compared with TernaryNet~\cite{heinrich2018ternarynet}, which was recently proposed for the compression of 2D U-Net~\cite{unet} for pancreas CT segmentation. Additionally, as an alternative compression method, we deployed knowledge distillation with a temperature $T=40$ to train scaled-down variants~\cite{hinton2015distilling} of 3D U-Net and V-Net,
that take up exactly the same storage space as the original-size models compressed with 3DQ.

\section{Results and Discussion}
\label{sec:results}

\begin{figure}[t]
\centering
  \includegraphics[width=0.88\textwidth]{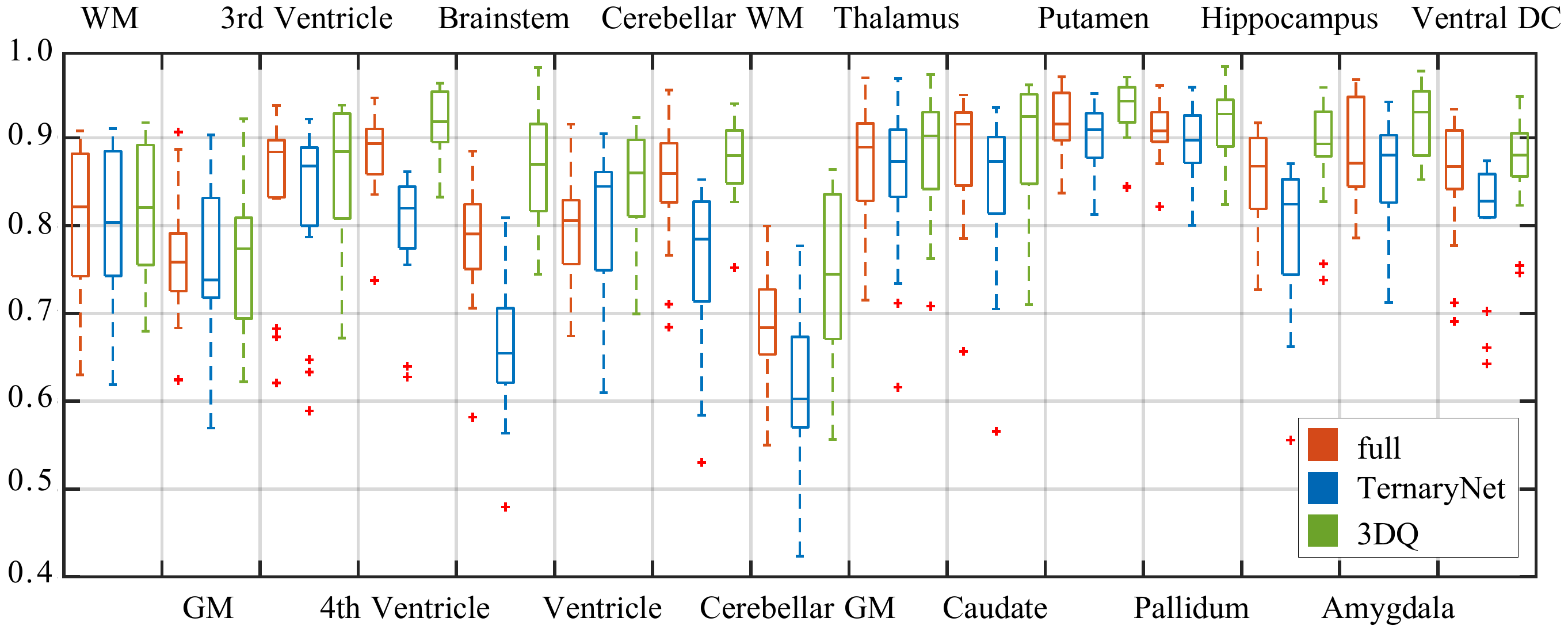}
   \caption{Box plot of Dice scores achieved by 3D U-Net comparing 3DQ with baseline methods on the right hemisphere classes of MALC.}
   \label{fig:boxplot}
\end{figure}

\noindent \textbf{Ablative Testing}
Table~\ref{tab:ablative} highlights that models quantized with ternary weights outperform their binary counterparts by 3-11\% for MALC and 7\% for HC, thanks to the higher learning capacity, motivating the choice of ternary weights in 3DQ.
Moreover, Table~\ref{tab:ablative} showcases the positive impact of the scaling factor $\alpha$ we introduced, allowing 3DQ to perform up to 1-2\% better than TTQ on both datasets,
with a lower standard deviation. $\alpha$ mitigates the quantization constraints, enabling the ternary weights to assume a larger variety of values, better resembling their full precision counterparts.

\begin{wraptable}{r}{6.7cm}
\vspace{-0.7cm}
\ra{1.3}
\resizebox{0.55\textwidth}{!}{
\begin{tabular}{@{}lccccccc@{}}%\toprule
\multicolumn{3}{c}{} & \multicolumn{1}{c}{BTQ} & \phantom{abc}& \multicolumn{1}{c}{TTQ} & \phantom{abc} & \multicolumn{1}{c}{3DQ}\\ \toprule
 \parbox[t]{2mm}{\multirow{1}{*}{\rotatebox[origin=c]{90}{\scriptsize{HC}}}} & & \tiny{3DU} &  0.847 $\pm$ 0.009 & & 0.912 $\pm$ 0.008 & & \textbf{0.915 $\pm$ 0.006} \\
\cmidrule{1-8}
 \parbox[t]{2mm}{\multirow{2}{*}{\rotatebox[origin=c]{90}{\scriptsize{MALC}}}} & & \tiny{V} & 0.770 $\pm$ 0.013 & & 0.790 $\pm$ 0.010 & & \textbf{0.802 $\pm$ 0.004} \\
%\cmidrule{3-11}
 & & \tiny{3DU} & 0.735 $\pm$ 0.005 & & 0.828 $\pm$ 0.007 & & \textbf{0.844 $\pm$ 0.006} \\
\bottomrule
\end{tabular}
}
\caption{Comparison of Dice scores of 3DQ with TTQ and its binarized version BTQ on HC and MALC, with 3D U-Net and V-Net.}
\label{tab:ablative}
\vspace{-0.4cm}
\end{wraptable}

\noindent \textbf{Comparative Methods}
As can be seen in Table~\ref{tab:baseline}, 3D U-Net quantized with 3DQ performs as good as the full precision model in the case of HC, while it outperforms it by over 2\% for MALC. This can be attributed to the quantization acting as a regularization technique by limiting the dynamic range of the weight values.
3DQ also outperformed TernaryNet across all experiments for MALC and HC, with a margin ranging from 7 to over 10\%, thanks to the learned scaling factors $\gamma_\pm$ and the absence of the hyperbolic ternary tangent which bounds the activation values and limits the learning capacity of the model.

Figure~\ref{fig:segmentation} shows sample segmentations for a slice of a MALC volume from 3D U-Net. A zoomed view of the segmentations, indicates important subcortical structures with a white arrow. Both full precision and TernaryNet predictions suffer from over-inclusion of small structures and spurious misclassified regions. The box plot in Figure~\ref{fig:boxplot} confirms the higher quality of the segmentations produced by 3DQ, reporting the Dice scores on the right hemisphere structures. 3DQ outperformed both full precision and TernaryNet, with fewer outliers, demonstrating more uniform results throughout the samples.

\noindent \textbf{Comparison with Knowledge Distillation}
Another experiment reported in Table~\ref{tab:baseline} is the comparison of 3DQ with knowledge distillation. In order to match the 3DQ model sizes while keeping the full precision weights, the distilled networks have 16x fewer parameters than the full models. Even though the smaller networks achieve almost equal performance to the full model for HC, the margin is increased for MALC, where the student networks achieved a 9-10\% worse Dice score than the full models for both 3D U-Net and V-Net. This drop in performance can be attributed to the fact that distilled models are 16x smaller than the original ones and additionally have to rely on the predictions of a teacher network, limiting their learning capacity. 3DQ also outperforms the distilled models across the board by a substantial 8-11\% on MALC, constituting a successful model compression choice.

\noindent \textbf{Quantization on Different Architectures}
Table~\ref{tab:baseline} highlights the effects of quantization in the two different 3D model architectures, specifically 3D U-Net and V-Net. Although 3D U-Net is 3x smaller than V-Net, it achieved higher Dice scores in our experiments on MALC, especially when quantized. While the full models performed similarly with a difference of 1\%, the quantized 3D U-Net achieved a 4\% higher Dice than the quantized V-Net. We attribute this difference in performance to the fact that MALC consists of 15 training volumes, which are considered limited data for V-Net, a large model with 45M trainable parameters, in contrast with 3D U-Net that has 16M.
\begin{table}[t]\centering\ra{1.3}
\resizebox{0.82\textwidth}{!}{
\begin{tabular}{@{}lccccccccc@{}}%\toprule
\multicolumn{3}{c}{} & \multicolumn{1}{c}{Full} & \phantom{abc}& \multicolumn{1}{c}{Distilled} & \phantom{abc}& \multicolumn{1}{c}{TernaryNet} & \phantom{abc} & \multicolumn{1}{c}{3DQ}\\\toprule
 \parbox[t]{2mm}{\multirow{1}{*}{\rotatebox[origin=c]{90}{\scriptsize{HC}}}} & & \tiny{3DU} &  0.914 $\pm$ 0.005 & & 0.908 $\pm$ 0.019 & & 0.845 $\pm$ 0.013 & & \textbf{0.915 $\pm$ 0.006} \\
\cmidrule{1-10}
 \parbox[t]{2mm}{\multirow{2}{*}{\rotatebox[origin=c]{90}{\scriptsize{MALC}}}} & & \tiny{V} & \textbf{0.815 $\pm$ 0.008} & & 0.715 $\pm$ 0.001 & & 0.696 $\pm$ 0.016 & & 0.802 $\pm$ 0.004 \\
%\cmidrule{3-11}
 & & \tiny{3DU} & 0.822 $\pm$ 0.005 & & 0.730 $\pm$ 0.008 & & 0.774 $\pm$ 0.012 & & \textbf{0.844 $\pm$ 0.006} \\
\bottomrule
\end{tabular}
}
\vspace{0.2cm}
\caption{Comparison of Dice scores of 3DQ with baseline methods. Tests performed on HC and MALC, with 3D U-Net and V-Net.}
\label{tab:baseline}
\vspace{-0.8cm}
\end{table}

\begin{wraptable}{r}{6.7cm}
\vspace{-0.95cm}
\ra{1.3}
\resizebox{0.55\textwidth}{!}{
\begin{tabular}{@{}lcccccccccc@{}}%\toprule
\multicolumn{2}{c}{} & \multicolumn{1}{c}{Full} & \phantom{abc}& \multicolumn{1}{c}{Distilled} & \phantom{abc}& \multicolumn{1}{c}{Ternary} & \phantom{abc} & \multicolumn{1}{c}{Binary} \\ \toprule
 3D U-Net & &  63MB & & 3.9MB & & 3.9MB & & \textbf{2.0MB} \\
 V-Net & & 175MB & & 11MB & & 11MB & & \textbf{5.5MB} \\
\bottomrule
\end{tabular}
}
\caption{Model size in MBytes for full precision and compressed models.}
\label{tab:compression}
\vspace{-0.7cm}
\end{wraptable} 

\noindent \textbf{Compression}
The models storage requirements are shown in Table~\ref{tab:compression}. By using ternary weights, TernaryNet, TTQ and 3DQ reduce the storage size by a factor of 16, compared to a full precision model. The different scaling factors impact the storage by only a few KBytes. Binary weights additionally reduce the storage required by 2x in comparison to ternary ones, at the cost of lower Dice scores, due to limited learning capacity.

\section{Conclusion}
\label{sec:conclusion}
In this paper we proposed 3DQ, a ternary quantization method, which was applied for the first time to 3D F-CNNs on the challenging task of volumetric whole brain segmentation. The models quantized with 3DQ achieved equal or better Dice scores than the baselines, including the full precision models, across two datasets. Thanks to 16x model compression, 3DQ constitutes a valid approach for storage-critical applications, as patient-specific networks or weights transferring during Federated Learning.

\bibliographystyle{unsrt}
\bibliography{refs.bib}
\end{document}